\documentclass[10pt,twocolumn,letterpaper]{article}

\usepackage{cvpr}              

\usepackage{graphicx}
\usepackage{amsmath}
\usepackage{amssymb}
\usepackage{booktabs}
\usepackage{pifont}
\newcommand{\cmark}{\ding{51}}%

\usepackage[accsupp]{axessibility}

\usepackage[pagebackref,breaklinks,colorlinks]{hyperref}

\usepackage[capitalize]{cleveref}
\crefname{section}{Sec.}{Secs.}
\Crefname{section}{Section}{Sections}
\Crefname{table}{Table}{Tables}
\crefname{table}{Tab.}{Tabs.}

\def\cvprPaperID{9295} 
\def\confName{CVPR}
\def\confYear{2022}

\begin{document}


\title{DirecFormer: A Directed Attention in Transformer Approach to \\ Robust Action Recognition\vspace{-4mm}}


\author{Thanh-Dat Truong$^{1}$, Quoc-Huy Bui$^{2}$, Chi Nhan Duong$^{3}$, Han-Seok Seo$^{4}$ \\Son Lam Phung$^{5}$, Xin Li$^{6}$, Khoa Luu$^{1}$\\
$^{1}$CVIU Lab, University of Arkansas \quad
$^{2}$NextG, FPT Software \\
$^{3}$Concordia University \quad
$^{4}$Dep. of Food Science, University of Arkansas \\
$^{5}$University of Wollongong \quad
$^{6}$West Virginia University\\
\tt\small \{tt032, khoaluu,  hanseok\}@uark.edu, bqhuy@apcs.fitus.edu.vn \\ 
\tt\small dcnhan@ieee.org, phung@uow.edu.au, Xin.Li@mail.wvu.edu
\vspace{-2mm}
}

\maketitle

\begin{abstract}
Human action recognition has recently become one of the popular research topics in the computer vision community. Various 3D-CNN based methods have been presented to tackle both the spatial and temporal dimensions in the task of video action recognition with competitive results. However, these methods have suffered some fundamental limitations such as lack of robustness and generalization, e.g., how does the temporal ordering of video frames affect the recognition results?
This work presents a novel end-to-end Transformer-based Directed Attention (DirecFormer) framework\footnote{The implementation of DirecFormer is available at \url{https://github.com/uark-cviu/DirecFormer}} for robust action recognition. The method takes a simple but novel perspective of Transformer-based approach to understand the right order of sequence actions.  Therefore, the contributions of this work are three-fold. Firstly, we introduce the problem of ordered temporal learning issues to the action recognition problem. Secondly, a new Directed Attention mechanism is introduced to understand and provide attentions to human actions in the right order. Thirdly, we introduce the conditional dependency in action sequence modeling that includes orders and classes. The proposed approach consistently achieves the state-of-the-art (SOTA) results compared with the recent action recognition methods \cite{timesformer,pan,vidtr,x3d}, on three standard large-scale benchmarks, i.e. Jester, Kinetics-400 and Something-Something-V2.

\end{abstract}

\section{Introduction}
\label{sec:intro}

\begin{figure}[!t]
    \centering
    \includegraphics[width=0.47\textwidth]{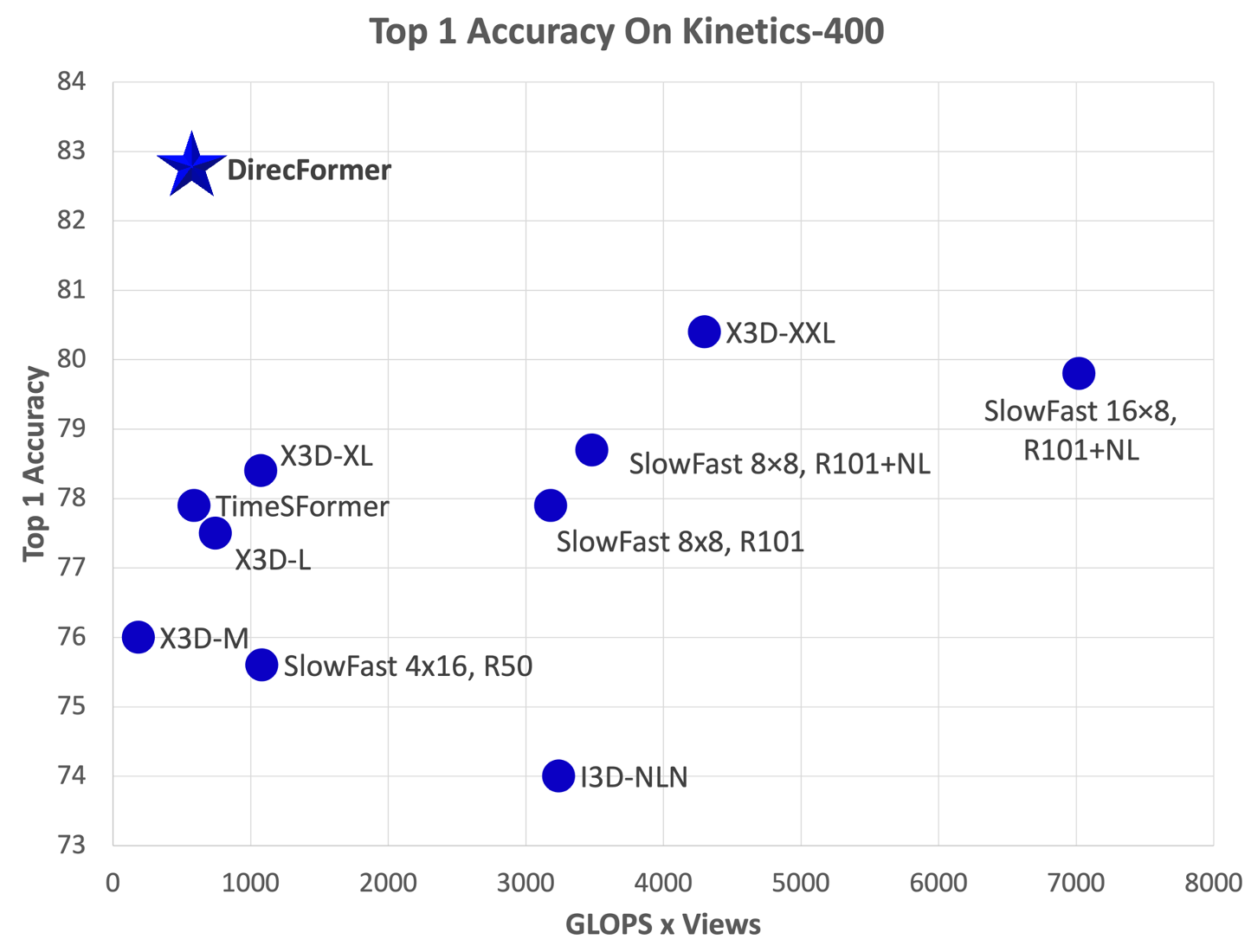}
    \caption{\textbf{Result Preview.} Top 1 accuracy against GLOPS $\times$ Views of our DirecFormer compared to other methods. DirecFormer achieves SOTA performance while maintaining the low computational cost.}
    \vspace{-0.2in}
    \label{fig:my_label1}
\end{figure}

Video understanding has recently become one of the popular research topics in the computer vision community. Video data has become ubiquitous and occurs in numerous daily activities and applications, e.g., movies and camera surveillance \cite{Duong_2017_ICCV, le2018segmentation, truong2021fastflow, duong2029cvpr_automatic}. 
In the field of video understanding \cite{dat2021bimal_iccv}, action recognition has become a fundamental problem. 
In action recognition, there is a need to pay more attention to the temporal structures of the video sequences. Indeed, emphasis on temporal modeling is a common strategy among most methods. It can be considered as the main difference between video and images. These works include long-short term dependencies \cite{duong2019learning, duong2019dam_ijcv}, temporal structure, low-level motion, and action modeling as a sequence of events or states.

The current methods in video action recognition utilize 3D or pseudo 3D convolution to extract the spatio-temporal features \cite{c3d, i3d, s3d, p3d}. However, these 3D CNN-based methods suffer from intensive computation with many parameters to be learned. Others attempt to adopt two-stream structures \cite{2streams_simonyan, 2streams_feichtenhofer, spatiotemporal_resnet, slowfast} for accurate action recognition, since information from one branch could be fused to the other one. Some methods in this category require computing the optical flow first, which could be time-consuming and requires a large amount of storage. Others apply 3D convolution to avoid computing the optical flow. Nonetheless, this approach also requires a large amount of computational resources to implement.

Although prior methods \cite{crosstransformer,noframe,actionnet,vatt,tdn} have achieved remarkable performance, they have several limitations related to the robustness of the models. In this paper, we therefore address two fundamental questions for current action recognition models.
In the first question, given a set of video frames \textit{shuffled in a random order} and different from the original one, will it be classified as the same label as the original recognition result?
If it is the case, these models have been clearly overfitted or biased to other factors (e.g. scene background), rather than learned semantic information of the actions. 
In the second question, we want to understand whether these action recognition models are able to \textit{correct the incorrectly-ordered frames} to the right ones and provide an accurate prediction? Finally, we introduce a new theory to improve the robustness and generalization of the action recognition models.

\subsection{Contributions of this Work}

In this work, we present a new end-to-end Transformer-based Directed Attention (DirecFormer) approach to robust action recognition. Our method takes a simple but novel perspective of Transformer-based approach to learn the right order of a sequence of actions. The contributions of this work are three-fold. 
First, we introduce the problem of ordered temporal learning in action recognition. Second, a new Directed Attention mechanism is introduced to provide human action attentions in the right order. Third, we introduce the conditional dependency in action sequence modeling that includes orders and classes. The proposed approach consistently achieves the State-of-the-Art results compared to the recent methods \cite{swin,vivit,pan} on three standard action recognition benchmarks, i.e. Jester \cite{jester}, Something-in-Something V2 \cite{ssv2} and Kinetics-400 \cite{kinetic400}, as in Fig. \ref{fig:my_label1}.

\begin{figure*}
    \centering
    \includegraphics[width=0.85\textwidth]{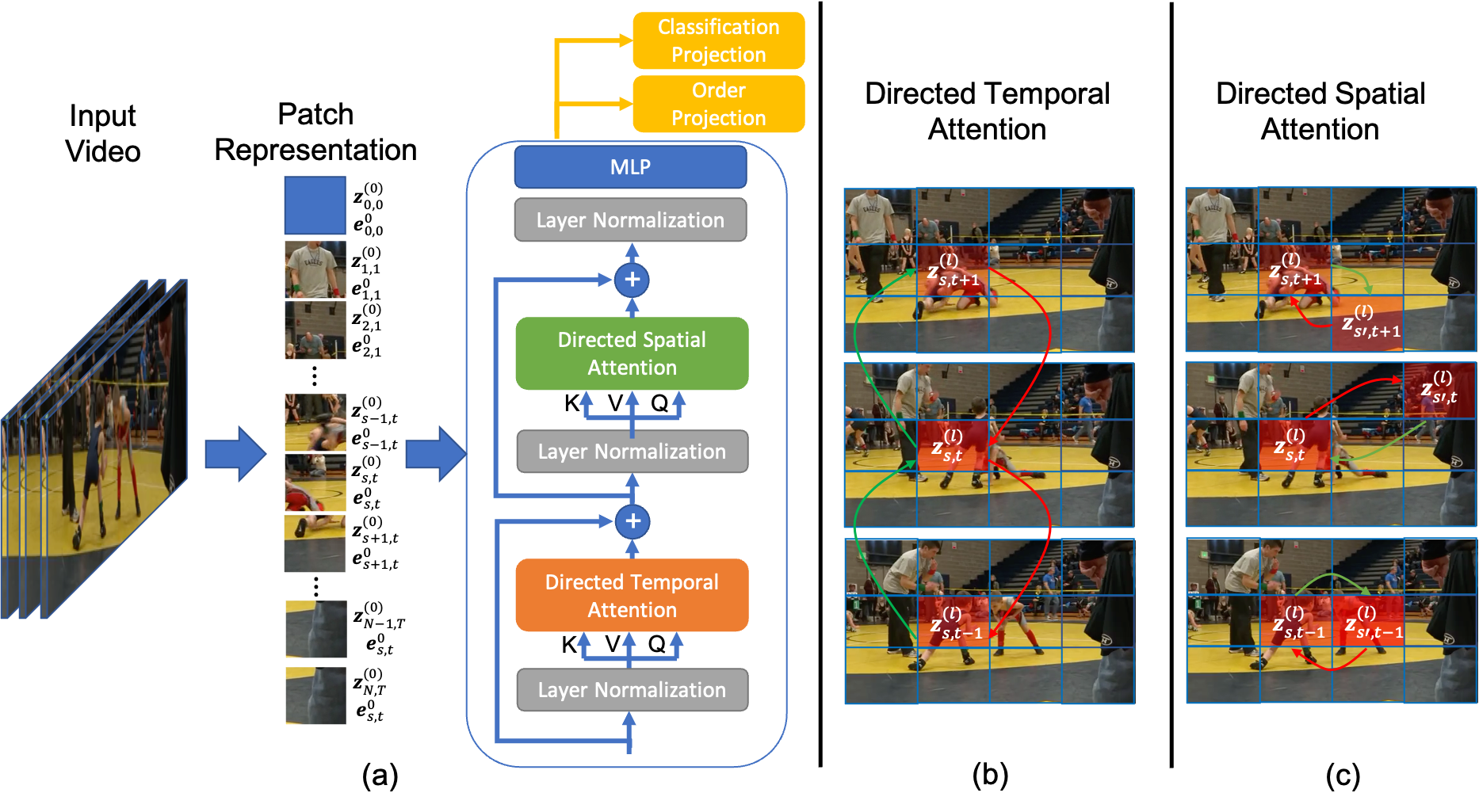}
    \vspace{-4mm}
    \caption{\textbf{The Proposed Framework.} (a) The Proposed DirecFormer. (b) Directed Temporal Attention. (c) Directed Spatial Attention. The green arrows in (b) and (c) denotes for positive correlation and the red arrows denotes for negative correlation.}
    \vspace{-0.2in}
    \label{fig:framework}
\end{figure*}

\section{Related Work}

\noindent
\textbf{Video Action Classification.} In recent years, video understanding has become a popular topic in Computer Vision due to its promising applications such as robotics, autonomous driving, camera surveillance or human behavior analysis. In the early days, many traditional approaches used hand-crafted features as a method to encode information of video sequences \cite{ivan_laptev, poitr_dollar, geert_willems, cordelia_schimid, invariant_spatiotemporal_feat, jason_corso, iDT}. Among these approaches, iDT \cite{iDT} achieved very good performance by utilizing dense trajectory features and became one of the most popular hand-designed methods.

Later, 
with the success of deep learning architecture 
\cite{alex_krizhevsky, vgg, szegedy, inceptionv3, resnet} using computing hardware, i.e., GPU, TPU, and the introduction of various large-scale datasets, e.g. Sport1M \cite{sport1m}, Kinetics \cite{kinetic400, kinetic600} and AVA \cite{ava}, 
video understanding, especially video action recognition, has become easier to approach in the research community, resulting in an introduction of a series of deep learning frameworks.
These methods mainly focus on learning spatio-temporal representations in an end-to-end classification manner. \cite{lrcn} proposed to model the temporal relationship using LSTM \cite{lstm} to corporate 2D CNN features. \cite{sport1m} presented an approach that fuses the information from the temporal dimension while suggesting applying a single 2D CNN to each frame of the video sequence. However, this method cannot handle well the motion change and perform weaker than the hand-crafted features methods. 

The remaining approaches for video action recognition could be divided into two categories. The first group contains models that adopt the conventional two-stream structure \cite{2streams_simonyan} to improve the temporal modeling capability. A spatial 2D CNN is used to learn semantic features and a temporal 2D CNN is applied in the other branch to analyze the motion content of the video sequences using the optical flow as input. Both streams are trained in parallel and the scores are averaged to make the final predictions. \cite{2streams_feichtenhofer, spatiotemporal_resnet, spatiotemporal_multiplier} studied different combinations to fuse spatio-temporal information between both streams. TSN \cite{tsn} proposed sampling sparse frames from evenly divided segments of the video clip to capture long-range dependencies. These dual-path methods require additional computation of the optical flow, which is time-consuming and demand a considerable amount of storage. However, our proposed method can operate without the need of optical flow modality, thus, reducing the complexity of the network.

The second category for action recognition is 3D CNN based methods and the (2+1)D CNN variants. C3D \cite{c3d} was the first work to apply 3D convolutions to model the spatial and temporal features together. I3D \cite{i3d} was proposed to inflate 2D convolutional kernels into 3D to capture spatio-temporal features. However, the major drawback of 3D CNNs is the large number of parameters involved. To cope with the intensive computation of 3D CNN, various methods adopted the 2D + 1D paradigm. P3D \cite{p3d} decomposes 3D convolution into a pseudo-3D convolutional block. R(2+1)D \cite{r21d} and S3D-G \cite{s3d} factorize the 3D convolution to enhance accuracy and reduce the complexity. TRN \cite{trn} introduced an interpretable relational module to replace the average pooling operation. TSM \cite{tsm} shifts part of the features forward and backward along the temporal dimension, allowing the network to achieve the performance of 3D CNN but maintains the complexity of 2D CNN. Non-local neural network \cite{non_local} proposed a special non-local operation for better capturing the long-range temporal dependencies between video frames. SlowFast \cite{slowfast} adopted a dual-path network to model the spatio-temporal information at two different temporal rates, with mid-level features being interactively fused. In general, our method also learns to approximate the spatio-temporal representation at the feature level with the help of the knowledge distillation process.

More recently, significant improvement in terms of efficiency has been reported for action recognition in \cite{chen2021deep}; it was found that 2D-CNN and 3DCNN models behave similarly in terms of spatio-temporal representation ability and transferability. More efficient action recognition can be achieved by focusing more on making the most of selected frames by dynamic knowledge propagation \cite{kim2021efficient} or exploiting spatio-temporal self-similarity \cite{kwon2021learning}. Most recent works such as elastic semantic network (Else-Net) \cite{li2021else} and memory attention network (MAN) \cite{li2021memory} also reported promising improvement in terms of recognition accuracy.

\noindent
\textbf{Video Ordering} Several prior works \cite{paper_A, paper_B, paper_C} have considered the frame orders into account.
Although these prior works have partially addressed some aspects of order prediction, their losses only provide a weak supervision, i.e. binary label for in- or out-of-order  \cite{paper_A, paper_C} or sub-clip based order \cite{paper_B}. Moreover, there is no explicit mechanism to enforce the model focus on the motion information rather than particular background information of the scene.

\noindent
\textbf{Video Transformer} Transformer approaches have filled an important role by acquiring competitive accuracy while maintaining computational resources compared with the traditional convolution method. A pure-transformer  based model (ViViT) was demonstrated in \cite{vivit} handling spatio-temporal tokens from a long sequence of frames by factorizing space-time dimension inputs efficiently on both large and small datasets. The divided spatial and temporal attention within each block, TimeSformer\cite{timesformer} reduces training time while achieving comparable test efficiency. A Spatial-Temporal Transformer network (ST-TR) was developed in \cite{plizzari2021spatial,zhang2021stst} for skeleton-based action recognition. A transformer-based RGB-D egocentric action recognition framework called Trear was proposed  in \cite{li2021trear} showing dramatic improvement over the existing state-of-the-art results. A multiscale pyramid network called MViT was proposed in \cite{mvit} to extract information from low-level to high levels of attention. When compared with many other successful applications of transformers, their potential in action recognition has still largely remained unexplored.

\section{The Proposed Method}

Let $\mathbf{x} \in \mathbb{R}^{T \times H \times W \times 3}$ be the input video and $\mathbf{y}$ be the corresponding label of the video $\mathbf{x}$. $H, W$ and $T$ are the height, the width and the number of frames of a video, respectively. Let $\mathbf{o} \in \mathbb{N}^{T}$ be the permutation representing the reordering of video frames and $\mathbf{i}$ be the indexing associated with the permutation. Our goal is to learn a deep network to classify the actions and infer the permutation simultaneously as in Eqn. \eqref{eq:MaxE}.
\begin{equation} \label{eq:MaxE}
\begin{split}
  \arg\max_{\theta} \mathbb{E}_{\mathbf{x, y, o, i}} \left(\log(p(\mathbf{y} | \mathbf{x}; \theta)) + \log(p(\mathbf{i} | \mathcal{T}(\mathbf{x}, \mathbf{o}); \theta))\right)
\end{split}
\end{equation}
where $\theta$ is the parameters of the deep neural network, 
and $\mathcal{T}$ is the permutation function. Given a video $\mathbf{x}$ and the permutation $\mathbf{o}$, the goal is to learn the class label $\mathbf{y}$ of the ordered video and learn the ordering $\mathbf{i}$ of the video after permutation $\mathcal{T}\mathbf{(x, o)}$.

To effectively predict the class label $\mathbf{y}$ and the indexing of the permutation $\mathbf{i}$, a Transformer with Directed Attention is introduced to learn the directed attention in both spatial and temporal dimensions. The proposed DirecFormer is therefore formulated as in Eqn. \eqref{eq:direcformer}.
\begin{equation} \label{eq:direcformer}
\begin{split}
    \mathbf{\hat{y}} &= \phi_{cls} \odot \mathcal{G} (\mathbf{x}) \\
    \mathbf{\hat{i}} &= \phi_{ord} \odot \mathcal{G} (\mathcal{T}(\mathbf{x}, \mathbf{o}))
\end{split}
\end{equation}
where $\mathcal{G}$ is the proposed DirecFormer; $\phi_{cls}$ and $\phi_{ord}$ are the projections that map the token outputted from DirecFormer to the predicted class label $\mathbf{\hat{y}}$ and the predicted ordering index $\mathbf{\hat{i}}$, respectively; and $\odot$ is the functional composition. Fig. \ref{fig:framework} illustrates our proposed framework.
The proposed DirecFormer method will be described in detail in the following section. 

\subsection{Patch Representation}

Given a video frame, it is represented by $N$ non-overlapped patches of $P \times P$ ($N = \frac{HW}{P^2}$) as in \cite{vit}. 
Let us denote $\mathbf{x}_{s, t} \in \mathbb{R}^{3P^2}$ as a vector representing the patch $s$ of the video frame $t$, where $s$ ($1 \leq s \leq N$) denotes the spatial position and $t$ represents the temporal dimension ($1 \leq t \leq T$).
To embed the temporal information into the representation, the raw patch representation is projected to the latent space with additive temporal representation as in Eqn. \eqref{eq:latent}.
\begin{equation} \label{eq:latent}
    \mathbf{z}^{(0)}_{s,t} = \alpha(\mathbf{x}_{s,t}) + \mathbf{e}_{s, t}
\end{equation}
where $\alpha$ is the embedding network and $\mathbf{e}_{s, t}$ is the spatial-temporal embedding added into the patch representation. The output sequences $\{\mathbf{z}^{(0)}_{s,t}\}_{s=1,t=1}^{N, T}$ represent the input tokens fed to our DirecFormer network. We also add one more learnable token $\mathbf{z}_{0,0}$ in the first position, as in BERT \cite{bert} to represent the classification token.

\subsection{Directed Attention Approach}

The proposed DirecFormer consists of $L$ encoding blocks. In particular, the current block $l$ takes the output tokens of the previous block $l-1$ as the input and decomposes the token into the key $\mathbf{k}^{(l)}_{s,t}$, value $\mathbf{v}^{(l)}_{s,t}$, and query $\mathbf{q}^{(l)}_{s,t}$ vectors as in Eqn. \eqref{eqn:key_value_query}.
\begin{equation} \label{eqn:key_value_query}
\begin{split}
    \mathbf{k}^{(l)}_{s,t} &= \beta^{(l)}_k\left(\tau^{(l)}_k\left(\mathbf{z}^{(l-1)}_{s,t}\right)\right) \\
    \mathbf{v}^{(l)}_{s,t} &= \beta^{(l)}_v\left(\tau^{(l)}_v\left(\mathbf{z}^{(l-1)}_{s,t}\right)\right) \\
    \mathbf{q}^{(l)}_{s,t} &= \beta^{(l)}_q\left(\tau^{(l)}_q\left(\mathbf{z}^{(l-1)}_{s,t}\right)\right) \\
\end{split}
\end{equation}
where $\beta^{(l)}_{k}$, $\beta^{(l)}_{v}$ and $\beta^{(l)}_{q}$ represent the key, value, and query embedding, respectively; $\tau^{(l)}_k$, $\tau^{(l)}_v$ and $\tau^{(l)}_q$ are the layer normalization \cite{layer_norm}. 

In the traditional self-attention approach, the attention matrix is computed by the scaled dot multiplication between key and query vectors. 
Although scaled dot attention has shown its potential performance in video classification, this attention is non-directed because it is unable to illustrate the direction of attention. 
In particular, the scaled dot attention simply indicates the correlations among tokens and ignores the temporal or spatial ordering among tokens. 
It is noticed that the ordering of frames in a video sequence does matter. The recognition of actions in a video is highly dependent on the ordering of video frames. For example,
the same group of video frames,  if ordered differently in time, may result in different actions, e.g.  walking might become running. 
However, traditional Softmax attention can not fully exploit the ordering of video frames because it does not contain the directional information of the correlation. 

Therefore, we propose a new Directed Attention using the cosine similarity. Formally, the attention weights $\mathbf{a}^{(l)}_{(s,t)}$ for a query $\mathbf{q}^{(l)}_{s,t}$ can be formulated as in Eqn. \eqref{eq:DAeqn}.
\begin{equation} \label{eq:DAeqn}
\footnotesize
\begin{split}
    \mathbf{a}^{(l)}_{(s,t)} = \left[\operatorname{cos}\left(\frac{\mathbf{q}^{(l)}_{s,t}}{\sqrt{D}}, \mathbf{k}^{(l)}_{0,0}\right) \; \left\{\operatorname{cos}\left(\frac{\mathbf{q}^{(l)}_{s,t}}{\sqrt{D}}, \mathbf{k}^{(l)}_{s',t'}\right)\right\}_{s'=1,t'=1}^{N,T}\right]
\end{split}
\end{equation}
where $D$ is the dimensional length of the query vector $\mathbf{q}^{(l)}_{s,t}$, $\mathbf{a}^{(l)}_{p,t} \in \mathbb{R}^{NT+1}$ denotes the directed attention weights. 
This attention is computed over the spatial and temporal dimensions. As a result, this operator suffers a heavy computational cost.
We therefore divide and conquer the Directed Attention in the spatial dimension and temporal dimension sequentially as in \cite{timesformer}. 

More specifically, we first implement the attention mechanism over the time dimension ($\mathbf{a}^{(l)-time}_{(s,t)}$) as in Eqn. \eqref{eq:AttTime}.
\begin{equation} \label{eq:AttTime}
\footnotesize
\begin{split}
    \mathbf{a}^{(l)-time}_{(s,t)} = \left[\operatorname{cos}\left(\frac{\mathbf{q}^{(l)}_{s,t}}{\sqrt{D}}, \mathbf{k}^{(l)}_{0,0}\right) \; \left\{\operatorname{cos}\left(\frac{\mathbf{q}^{(l)}_{s,t}}{\sqrt{D}}, \mathbf{k}^{(l)}_{s,t'}\right)\right\}_{t'=1}^{T}\right]
\end{split}
\end{equation}
Then, the directed temporal attention information is accumulated to the current token representations as in Eqn. \eqref{eqn:time_att_vec}.
\begin{equation} \label{eqn:time_att_vec}
\begin{split}
    \mathbf{s}^{(l)-time}_{s,t} &= \mathbf{a}^{(l)-time}_{(s, t), (0, 0)}\mathbf{v}^{(l)}_{0,0} + \sum_{t'=1}^T \mathbf{a}^{(l)-time}_{(s, t), (s, t')}\mathbf{v}^{(l)}_{s,t'} \\
    \mathbf{z'}^{(l)-time}_{s,t} &= \mathbf{z}^{(l-1)}_{s,t} +  \gamma^{(l)-time}\left(\mathbf{s}^{(l)-time}_{s,t}\right) 
\end{split}
\end{equation}
where $\gamma^{(l)-time}$ denotes the temporal projection. Secondly, the temporally attentive vector $\mathbf{z'}^{(l)-time}_{s,t}$ is projected to the new key, value, and query to drive Spatial Directed Attention as in Eqn. \eqref{Eq:SDA2}.
\begin{equation} \label{Eq:SDA2}
\begin{split}
    \mathbf{k'}^{(l)}_{s,t} &= \beta'^{(l)}_k\left(\tau'^{(l)}_k\left(\mathbf{z'}^{(l)-time}_{s,t}\right)\right) \\
    \mathbf{v'}^{(l)}_{s,t} &= \beta'^{(l)}_v\left(\tau'^{(l)}_v\left(\mathbf{z'}^{(l)-time}_{s,t}\right)\right) \\
    \mathbf{q'}^{(l)}_{s,t} &= \beta'^{(l)}_q\left(\tau'^{(l)}_q\left(\mathbf{z'}^{(l)-time}_{s,t}\right)\right) \\
\end{split}
\end{equation}
Next, the Directed Attention over the spatial dimension ($\mathbf{a}^{(l)-space}_{(s,t)}$) can be computed as in Eqn. \eqref{eq:DAtt3}.
\begin{equation} \label{eq:DAtt3}
\footnotesize
\begin{split}
    \mathbf{a}^{(l)-space}_{(s,t)} = \left[\operatorname{cos}\left(\frac{\mathbf{q'}^{(l)}_{s,t}}{\sqrt{D}}, \mathbf{k'}^{(l)}_{0,0}\right) \; \left\{\operatorname{cos}\left(\frac{\mathbf{q'}^{(l)}_{s,t}}{\sqrt{D}}, \mathbf{k'}^{(l)}_{s,t'}\right)\right\}_{t'=1}^{T}\right]
\end{split}
\end{equation}
The Spatial Directed Attention is then embedded to the temporal attentive features $\mathbf{z'}^{(l)-time}_{s,t}$ to obtain a new spatial attentive feature $\mathbf{z'}^{(l)-space}_{s,t}$ as in Eqn. \eqref{eq:SDAtt3}.
\begin{equation} \label{eq:SDAtt3}
\begin{split}
    \mathbf{s}^{(l)-space}_{s,t} &= \mathbf{a}^{(l)-space}_{(s, t), (0, 0)}\mathbf{v}^{(l)}_{0,0} + \sum_{s'=1}^N \mathbf{a}^{(l)-space}_{(s, t), (s', t)}\mathbf{v}^{(l)}_{s',t} \\
    \mathbf{z'}^{(l)-space}_{s,t} &= \mathbf{z'}^{(l)-times}_{s,t} +  \gamma^{(l)-space}\left(\mathbf{s}^{(l)-space}_{s,t}\right)  \\
\end{split}
\end{equation}
Finally, the Spatial-Temporal Attentive features $\mathbf{z'}^{(l)-space}_{s,t}$ are projected to the output token, getting ready for the next transformer block.

Formally, the output of the current transformer block ($\mathbf{z}^{(l)}_{s,t}$) can be formed as in Eqn. \eqref{eq:transblock}.
\begin{equation} \label{eq:transblock}
\begin{split}
    \mathbf{z}^{(l)}_{s,t} &= \varphi^{(l)}\left(\tau^{(l)}\left(\mathbf{z'}^{(l)-space}_{s,t}\right)\right) + \mathbf{z'}^{(l)-space}_{s,t}
\end{split}
\end{equation}
where $\varphi^{(l)}$ is a projection mapping implemented using a multi-layer perception network, and $\tau^{(l)}$ denotes the layer normalization \cite{layer_norm}.

\subsection{Classification Embedding}

The final representation is obtained in the final block of DirecFormer. Then, the class index and the order index of the video are predicted using linear projections as follows:
\begin{equation}
\begin{split}
    \mathbf{\hat{y}} = \phi_{cls}\left(\tau_{cls}\left(\mathbf{z}^{(L)}_{0,0}\right)\right) \\
    \mathbf{\hat{i}} = \phi_{odr}\left(\tau_{odr}\left(\mathbf{z}^{(L)}_{0,0}\right)\right) 
\end{split}
\end{equation}
where $\phi_{cls}$ and $\phi_{ord}$ are the classification projection and order projection, respectively; $\tau_{cls}$ and $\tau_{ord}$ are the layer normalization \cite{layer_norm}.

\subsection{Self-supervised Guided Loss For Directed Temporal Attention Loss}

In this stage, we are given the permutation of the current input video. To further reduce the burden of the network when learning the temporal attention, we propose a new self-supervised guided loss to enforce the temporal attention learning from the prior order knowledge. Formally, the self-supervised loss can be formulated as in Eqn. \eqref{eq:newloss}.
\begin{equation} \label{eq:newloss}
\small
    \mathcal{L}_{self} = \frac{1}{LNT^2} \sum_{l=1}^{L}\sum_{s=1,t=1}^{N,T}\sum_{t'=1}^{T}\left(1 - \mathbf{a}^{(l)-time}_{(s,t),(s,t')}\right) \varsigma(\mathbf{o}_t, \mathbf{o}_{t'})
\end{equation}
where $\varsigma(\mathbf{o}_t, \mathbf{o}_{t'}) = 1$ if the index $\mathbf{o}_t < \mathbf{o}_{t'}$, otherwise $\varsigma(\mathbf{o}_t, \mathbf{o}_{t'}) = -1$. The guided loss $\mathcal{L}_{self}$ helps to indicate the attention learning the correct direction during the training process.
Finally, the total loss function of DirecFormer is defined as in Eqn. \eqref{eq:finalloss}.
\begin{equation} \label{eq:finalloss}
    \mathcal{L} = \lambda_{cls}\mathcal{L}_{cls} + \lambda_{ord}\mathcal{L}_{ord} + \lambda_{self}\mathcal{L}_{self}
\end{equation}
where $\mathcal{L}_{cls}$ and $\mathcal{L}_{ord}$ are the cross-entropy losses of the classification projection ($\phi_{cls}$) and order projection ($\phi_{ord}$), respectively;
$\{\lambda_{cls}, \lambda_{ord}, \lambda_{self}\}$ are the parameters controlling their relative importance.

\begin{figure}
    \centering
    \includegraphics[width=0.45\textwidth]{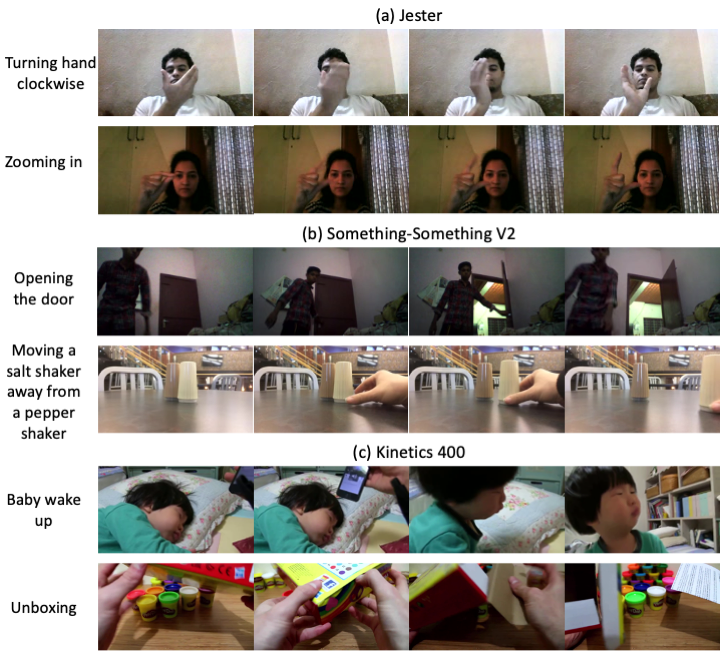}
    \vspace{-4mm}
    \caption{The Video Samples of Three Datasets: (a) Jester, (b) Something-Something V2, and (c) Kinetics-400.}
    \vspace{-0.2in}
    \label{fig:database}
\end{figure}

\section{Experiments}

In this section, we present the evaluation results with DirecFormer on three popular action recognition benchmarking datasets, i.e. Jester \cite{jester}, Something-Something V2 \cite{ssv2}, and Kinetics 400 \cite{kinetic400}. Firstly, we describe our implementation details and datasets used in our experiments. Secondly, we analyze our results with different settings shown in the ablation study on the Jester dataset. Lastly, we present our results on Something-Something V2 and Kinetics compared to prior state-of-the-art methods.

\subsection{Implementation Details}

The architecture of DirecFormer consists of $L=12$ blocks. The input video consists of $T=8$ frames sampled at a rate of $1/32$ and the resolution of each frame is $224 \times 224$ ($H = W = 224$).
The patch size is set to $18 \times 18$; therefore, there are $N = \frac{224^2}{16^2} = 196$ patches in total for each frame.
The embedding network $\alpha$ is implemented by a linear layer in which the output dimension is set to $768$. All values ($\beta^{(l)}_v, \beta'^{(l)}_v$), key ($\beta^{(l)}_k, \beta'^{(l)}_k$), query ($\beta^{(l)}_q, \beta'^{(l)}_q$) embedding networks, and projections ($\gamma^{(l)-time}, \gamma^{(l)-space}$) are also implemented by the linear layers.
Similar to \cite{vit,timesformer}, we adopt the multi-head attention in our implementation, where the number of heads is set to $12$. 
The network $\varphi^{(l)}$ is implemented as the residual-style multi-layer perceptron consisting of two fully
connected layers followed by a normalization layer.
Finally, the classification projection ($\phi_{cls}$) and the order projection ($\phi_{ord}$) are implemented as the linear layer.  We set the control parameters of loss to $1.0$, i.e. $\lambda_{cls} = \lambda_{ord} = \lambda_{self} = 1.0$. 

There will be a total of $T!$ permutations of the video frames. Therefore, learning with all permutations is ineffective. Moreover, the permutation set plays an important role. If these two permutations are very far from each other, the network may easily predict the order since the two permutations have significant differences.
However, if all the permutations are close to each other, learning the temporal attention is more challenging since the two permutations have minor differences in order. Therefore, we select 1,000 random permutations from $T! = 8!$ permutations so that the Hamming distance between permutations is as minimum as possible. Similar to \cite{domain_jigsaw_puzzle}, we use a greedy algorithm to generate the set of permutations.

In the evaluation, following the protocol of other papers \cite{slowfast, timesformer,x3d}, the single clip is sampled in the middle of the video. We use three spatial crops (top-left, center, and bottom-right) from the temporal clip and obtain the final result by averaging the prediction scores for these three crops.

\subsection{Datasets}

\noindent
\textbf{Jester.} \cite{jester} This dataset is a large-scale gesture recognition real-world video dataset that includes $148,092$ videos of $27$ actions. Each video is recorded for approximately 3 seconds. Fig \ref{fig:database}(a) illustrates the video samples of Jester. 

\noindent
\textbf{Something-in-Something V2.} \cite{ssv2} The dataset is a large-scale dataset to show humans performing predefined basic actions with everyday objects, which includes $174$ classes. It contains $220,847$ videos, with $168,913$ videos in the training set, $24,777$ videos in the validation set, and $27,157$ videos in the testing set. Similar to other work \cite{msnet,trg,timesformer,slowfastmg}, we report the accuracy on the validation set. Fig. \ref{fig:database}(b) shows two examples of two different class of Something-Something V2.
The licenses of Something-Something V2 and Jester datasets are registered by the TwentyBN team that are publicly available for academic research purposes.

\noindent
\textbf{Kinetics-400.} \cite{kinetic400} The dataset contains 400 human action classes, with at least 400 videos for each action. In particular, Kinetics-400 contains $234,619$ training videos and $19,761$ validation videos. 
The videos were downloaded from youtube and each video lasts for 10 seconds.
There are different types of human actions: \textit{Person Actions} (e.g. singing, smoking, sneezing, etc.); \textit{Person-Person Actions} (e.g. wrestling, hugging, shaking hands, etc.); and \textit{Person-Object Actions} (e.g. opening a bottle, walking the dog, using a computer, etc.).  
Fig. \ref{fig:database}(c) illustrates the video examples of Kinetics-400.
In our experiment, following the protocol of other papers \cite{slowfastmg,x3d,timesformer,x3d,slowfast}, we report the accuracy on the validation set.
The license of Kinetics is registered by Google Inc. under a Creative Commons Attribution 4.0 International License.

\subsection{Ablation Study}

\begin{table}[t]
    \centering
    \caption{\textbf{Ablation Study On Jester.} $X-Y$ denotes for the attention types of temporal and spatial dimension, respectively.  $X$ (and $Y$) could be either $S$: Softmax or $C$: Cosine.} 
    \resizebox{.47\textwidth}{!}{
    \begin{tabular}{c|c|c|c|c c}
        \textbf{Models}      & \begin{tabular}{@{}c@{}} \textbf{Attention}\\\textbf{Time-Space} \end{tabular}  & $\mathcal{L}_{ord}$ & $\mathcal{L}_{self}$& \textbf{Top 1} & \textbf{Top 5} \\ \hline
        I3D \cite{i3d}          & $-$   & $-$ & $-$ & 91.46 & 98.67 \\
        3D SqueezeNet \cite{squeezenet} & $-$   & $-$ & $-$ & 90.77 & $-$     \\ 
        ResNet 50 \cite{resnet}     & $-$   & $-$ & $-$ & 93.70 & $-$     \\ 
        ResNet 101 \cite{resnet}    & $-$   & $-$ & $-$ & 94.10 & $-$     \\ 
        ResNeXt \cite{resnext}       & $-$   & $-$ & $-$ & 94.89 & $-$     \\ 
        PAN \cite{pan}           & $-$   & $-$ & $-$ & 96.70 & $-$     \\ 
        STM \cite{stm}           & $-$   & $-$ & $-$ & 96.70 & $-$     \\ 
        ViViT-L/16x2 320 \cite{vivit} & $-$ & $-$ & $-$ & 81.70 & 93.80 \\
        TimeSFormer \cite{timesformer}   & $S-S$ & $-$ & $-$ & 94.14 & 99.19 \\ \hline
        \hline
        DirecFormer   & $S-C$ &  &  & 94.52 & 99.26 \\
        DirecFormer   & $S-C$ & \cmark &  & 94.65 & 99.25 \\
        \hline
        DirecFormer   & $C-S$ &  &  & 95.52 & 99.20 \\
        DirecFormer   & $C-S$ & \cmark &  & 96.28 & 99.45 \\
        DirecFormer   & $C-S$ & \cmark & \cmark & 97.55 & 97.54 \\ 
        \hline
        DirecFormer   & $C-C$ &  &  & 96.15 & 99.38 \\ 
        DirecFormer   & $C-C$ &  &  & 97.48 & 99.48 \\ 
        DirecFormer   & $C-C$ & \cmark & \cmark & 98.15 & 99.57  \\
    \end{tabular}
    }
    \label{tab:jester_exp}
    \vspace{-4mm}
\end{table}

\noindent
\textbf{Effectiveness Of Directed Attention}
To show the effectiveness of our proposed Directed Attention, we consider three different types of the temporal-spatial attention: (i) Softmax Temporal Attention followed by Cosine Spatial Attention (DirecFormer $S-C$), (ii) Cosine Temporal Attention followed by Softmax Spatial Attention (DirecFormer $C-S$), and (iii) Cosine Temporal Attention followed by Cosine Spatial Attention (DirecFormer $C-C$). The method is also compared with TimeSformer where the softmax attention is applied for both time and space. 
Table \ref{tab:jester_exp} illustrates the results of the DirecFormer with different settings compared to TimeSFormer and other approaches. In all configurations, our proposed DirecFormer outperforms the prior methods.

Considering the effectiveness of the directed attention in time and space,
the directions of the attention over the spatial dimension are important in some cases. For example, if $A$ performs an action to $B$ then $B$ receives an action from $A$. 
Considering the mentioned example, the spatial attention should involve directions so that the model can learn the actor(s) performing actions in a video.
However, the order of the temporal dimension plays a more important role in a video compared to the spatial dimension, since the order of the frames represents how the action is happening. 
As in Table \ref{tab:jester_exp}, the results of DirecFormer $C-S$ are better than DirecFormer $S-C$ confirming our hypothesis about the importance of time and space.
When the Directed Attention is deployed in both temporal and spatial dimensions, the results of DirecFormer $C-C$ were significantly improved and achieved the SOTA performance on the Jester dataset.

\noindent
\textbf{Effectiveness Of Losses}
With the order prediction loss $\mathcal{L}_{ord}$, the performance of the DirecFormer in all settings has been improved, since the prediction loss influences the way that network learns the Directed Temporal Attention. Moreover, the performance of DirecFormer is improved by employing the self-supervised guided loss $\mathcal{L}_{self}$. This self-supervised loss further enhances the directed temporal attention learning during the training. Consequently, the performance of DirecFormer is consistently improved by using our proposed losses, as in Table \ref{tab:jester_exp}.

\begin{table}[t]
    \small
    \centering
    \caption{\textbf{Order Correction By Hamilton Algorithm Performance On Jester.} $X-Y$ denotes for the attention types of temporal and spatial dimension, respectively.  $X$ (and $Y$) could be either $S$: Softmax or $C$: Cosine.} 
    \vspace{-3mm}
    \resizebox{.47\textwidth}{!}{
    \begin{tabular}{c|c|c|c|c}
        \textbf{Models}      & \begin{tabular}{@{}c@{}} \textbf{Attention}\\\textbf{Time-Space} \end{tabular}  & $\mathcal{L}_{ord}$ & $\mathcal{L}_{self}$& \textbf{OrderAcc} \\ \hline
        TimeSFormer \cite{timesformer}   & $S-S$ & $-$    & $-$ &  52.84 \\
        TimeSFormer \cite{timesformer}   & $S-S$ & \cmark & $-$ &  72.57 \\
        \hline
        \hline
        DirecFormer   & $C-S$ &  &  & 75.04 \\
        DirecFormer   & $C-S$ & \cmark &  & 87.16 \\
        DirecFormer   & $C-S$ & \cmark & \cmark & 90.02 \\
        \hline
        DirecFormer   & $C-C$ &  &  & 76.16 \\ 
        DirecFormer   & $C-C$ & \cmark &  & 88.96 \\ 
        \textbf{DirecFormer}   & $C-C$ & \cmark & \cmark & \textbf{90.19} \\ 
    \end{tabular}
    }
    \label{tab:order_exp}
    \vspace{-6mm}
\end{table}

\noindent
\textbf{Order Correction} To illustrate the ability of order learning of DirecFormer, we conduct an experiment in which, given a random temporal order video,  we show our approaches can retrieve back the correct order of the video from the directed temporal attention. 
In this experiment, we use the temporal attention of the last block and average this temporal attention over the spatial dimension. Then, we perform a search algorithm to find the Hamiltonian path on the temporal attention to find the correct order. 
In particular, we consider the temporal attention as the adjacency matrix of the graph, in which each frame is the node of the graph. The Hamilton path is the path that goes through each node exactly once (no revisit). Since our attention represents both direction and correlation among the frames, the higher (positive) correlation is, the higher the possibility of correct order should be between frames. Therefore, the Hamilton path with maximum total weight is going to represent the order of the video should be. 

Let $\mathbf{\hat{o}}$ be the order obtained by the Hamilton algorithm, the accuracy of the order retrieval can be defined as follows:
\vspace{-3mm}
\begin{equation}\label{eq:orderacc}
\small
    \operatorname{OrderAcc} = \frac{\operatorname{LCS}(\mathbf{\hat{o}}, \mathbf{o})}{T} \times 100
\end{equation}
\vspace{-1mm}
where $\operatorname{LCS}(\mathbf{\hat{o}}, \mathbf{o})$ is the longest common subsequence between $\mathbf{\hat{o}}$ and $\mathbf{o}$.
In this evaluation, for each video, we randomly select a permutation of $\{1,...,N\}$ as the order of the input video. To be fair between benchmarks, we set the same random seed value at the beginning of the testing script so that every time we conduct the evaluation, we obtain the same permutation for each video.

As shown in Table \ref{tab:order_exp}, we use the Softmax attention of TimeSFomer to retrieve the order of the video. The order accuracy of the TimeSFormer is only $52.84$. In other words, the Softmax attention of TimeSFormer can only predict the correct order of approximately 4 frames over 8 frames. With the support of order prediction loss, the order accuracy of TimeSFormer is improved to $72.57\%$. 
However, 
without the order prediction loss, our DirecFormer $C-S$ and DirecFormer $C-C$ have already correctly predicted the order of approximately 6 frames over 8 frames ($75.04\%$ and $76.16\%$).
When we further employ the order prediction and self-supervised guided losses, the performance of DirecFormer is significantly improved. Particularly, with the order prediction loss only, DirecFormer in all settings gains more than $87.0\%$ (which is approximately 7 frames over 8 frames). When both losses ($\mathcal{L}_{ord}$ and $\mathcal{L}_{self}$) are employed, the order accuracy of both DirecFormer $C-S$ and DirecFormer $C-C$ is improved to $90.02\%$ and $90.19\%$, respectively.
It should be noted that the performance of DirecFormer $C-C$ is only minorly greater than DirecFormer $C-S$ as the directed attention over the space does not largely affect the temporal order predictions.

\begin{table}[b]
    \footnotesize
    \centering
    \vspace{-4mm}
    \caption{\textbf{Comparison with the SOTA methods on Something-Something V2.} $X-Y$ denotes for the attention types of temporal and spatial dimension, respectively.  $X$ (and $Y$) could be either $S$: Softmax or $C$: Cosine.}
    \vspace{-4mm}
    \begin{tabular}{c | c | c c}
        \textbf{Models}         & \begin{tabular}{@{}c@{}} \textbf{Attention}\\\textbf{Time-Space} \end{tabular} &  \textbf{Top 1} & \textbf{Top 5} \\ \hline
        MSNet \cite{msnet}                   & $-$ & 63.00          & 88.40          \\ 
        SlowFast \cite{slowfast}                & $-$ & 63.00          & 88.50          \\ 
        SlowFast Multigrid \cite{slowfastmg}      & $-$ & 63.50          & 88.70 \\ 
        TRG \cite{trg}                     & $-$ & 62.20	       & \textbf{90.30}	 \\
        VidTr-L \cite{vidtr}                 & $-$ & 60.20           & $-$ \\
        TimeSFormer \cite{timesformer}             & $S-S$ & 59.10          & 85.60          \\ 
        TimeSFormer $-$ HR \cite{timesformer}     & $S-S$ & 61.80          & 86.90          \\ 
        TimeSFormer $-$ L \cite{timesformer}       & $S-S$ & 62.00          & 87.50          \\ 
        \hline \hline
        DirecFormer            & $S-C$ & 61.70          & 85.20     \\ 
        DirecFormer            & $C-S$ & 63.85          & 85.92          \\ 
        \textbf{DirecFormer}   & $C-C$ & \textbf{64.94} & 87.90          \\ 
    \end{tabular}
    \label{tab:ssv2_exp}
\end{table}

\subsection{Comparison with State-of-the-Art Results}

\noindent
\textbf{Something-Something V2}
Table \ref{tab:ssv2_exp} illustrates the performance of our proposed approaches evaluated on Something-Something V2 compared to prior SOTA approaches. In this experiment, similar to other approaches \cite{timesformer}, we use the DirecFormer pretrained on ImageNet-1K \cite{imagenet}.
As in Table \ref{tab:ssv2_exp}, our results in all settings outperform other candidates. 
With the simple design of the Transformer network with the directed attention mechanisms over time and space, our approaches achieve SOTA performance compared to traditional 3D CNN approaches  \cite{slowfast, msnet} and other Transformer approaches \cite{timesformer, vidtr} by a competitive margin.

\begin{table}[t]
\footnotesize
\centering
    \caption{\textbf{Comparison with the SOTA methods on Kinetics 400.} $X-Y$ denotes for the attention types of temporal and spatial dimension, respectively.  $X$ (and $Y$) could be either $S$: Softmax or $C$: Cosine.}
    \vspace{-5mm}
    \begin{tabular}{c | c | c c}
         \textbf{Models}         & \begin{tabular}{@{}c@{}} \textbf{Attention}\\\textbf{Time-Space} \end{tabular} &  \textbf{Top 1} & \textbf{Top 5} \\
        \hline
        I3D NLN \cite{i3d}            & $-$ & 74.00 & 91.10          \\
        ip-CSN-152 \cite{ipcsn152}         & $-$ & 77.80 & 92.80          \\
        LGD-3D-101 \cite{lgd3d}         & $-$ & 79.40 & 94.40          \\
        SlowFast \cite{slowfast}           & $-$ & 77.00 & 92.60   \\
        SlowFast Multigrid \cite{slowfastmg} & $-$ & 76.60 & 92.70   \\
        X3D-M \cite{x3d}              & $-$ & 75.10 & 91.70   \\
        X3D-L \cite{x3d}              & $-$ & 76.90 & 92.50   \\
        X3D-XXL \cite{x3d}            & $-$ & 80.40 & 94.60   \\
        MViT \cite{mvit}               & $-$ & 78.40 & 93.50    \\
        TimeSFormer \cite{timesformer}        & $S-S$ & 77.90 & 93.20 \\
        TimeSFormer $-$ HR \cite{timesformer} & $S-S$ & 79.70 & 94.40 \\
        TImeSFormer $-$ L \cite{timesformer}  & $S-S$ & 80.70 & 94.70 \\
        \hline
        \hline
        DirecFormer          & $S-C$ & 80.16 & 94.55 \\ 
        DirecFormer          & $C-S$ & 81.69 & 94.62 \\ 
        \textbf{DirecFormer} & $C-C$ & \textbf{82.75} & \textbf{94.86} \\ 
        \end{tabular}
        \vspace{-0.2in}
    \label{tab:k400_exp}
\end{table}

\noindent
\textbf{Kinetics 400} 
We conduct the experiments on Kinetics 400 and compare our results with prior SOTA methods.
The pretrained model on ImageNet-21K \cite{imagenet} for our DirecFormer is used, similar to \cite{timesformer}.
It is noted that the prior methods \cite{slowfast,x3d} use 10 temporal clips with 3 spatial crops of a video in the evaluation phase.
However, TimeSFormer and our DirecFormer use only 3 spatial crops of a video with a single clip to achieve the solid results.
In particular, our method achieves the SOTA performance compared to prior methods as shown in Table \ref{tab:k400_exp}. 
The Top 1 accuracy of the best model is approximately $2\%$ higher than TimeSFormer-L \cite{timesformer} sitting at $82.75\%$.
The effectiveness of the proposed directed attention has been also proved in these experiments, as the performance of DirecFormer is consistently improved when we deploy the directed attention over time and space.

\begin{table}[!b]
\footnotesize
\centering
\vspace{-3mm}
\caption{\textbf{Network Size Comparison.} We report the computational cost of the inference phase with a single ``view'' (temporal clip with spatial crop) $\times$ the numbers of such views used (GFLOPs $\times$ views). ``N\/A'' indicates the number is not available for us.}
\vspace{-2mm}
    \begin{tabular}{c|c|c}
        \textbf{Model}    & \textbf{GFLOPS x Views} & \textbf{Params} \\ 
        \hline
        I3D \cite{i3d}               & $108 \times $N\/A             & 12.0M             \\ \hline
        SlowFast 8x8 R50 \cite{slowfast}  & $36.1 \times 30$              & 34.4M           \\ \hline
        SlowFast 8x8 R101 \cite{slowfast} & $106 \times 30$               & 53.7M           \\ \hline
        Nonlocal R50 \cite{non_local}      & $282 \times 30$               & 35.3M           \\ \hline
        X3D-XL \cite{x3d}            & $35.8 \times 30$              & 11.0M           \\ \hline
        X3D-XXL \cite{x3d}           & $143.5 \times 30$             & 20.3M           \\ \hline
        ViViT-L/16x2 320 \cite{vivit}           & $3980 \times 3$ &  310.8M \\ \hline
        TimeSformer \cite{timesformer}       & $196 \times 3$                & 121.4M          \\ \hline
        DirecFormer       & $196 \times 3$                & 121.4M          \\ 
        \end{tabular}
    \label{tab:networksize}
\end{table}

\noindent
\textbf{Network Size Comparison} As shown in Table \ref{tab:networksize},
 although the number of parameters and the GFLOPS of single view used in our DirecFormer are higher than the traditional 3D-CNN approaches \cite{i3d, slowfast, x3d}, we only use 3 views compared to 30 views of prior approaches and maintain competitive performance. 
In comparison with TimeSFormer, we gain the same performance in terms of network size and inference flops; however, we achieve better accuracy on three large-scale benchmarks as shown in Tables \ref{tab:jester_exp}, \ref{tab:ssv2_exp}, and \ref{tab:k400_exp}.

\begin{figure}[!t]
    \centering
    \includegraphics[width=0.45\textwidth]{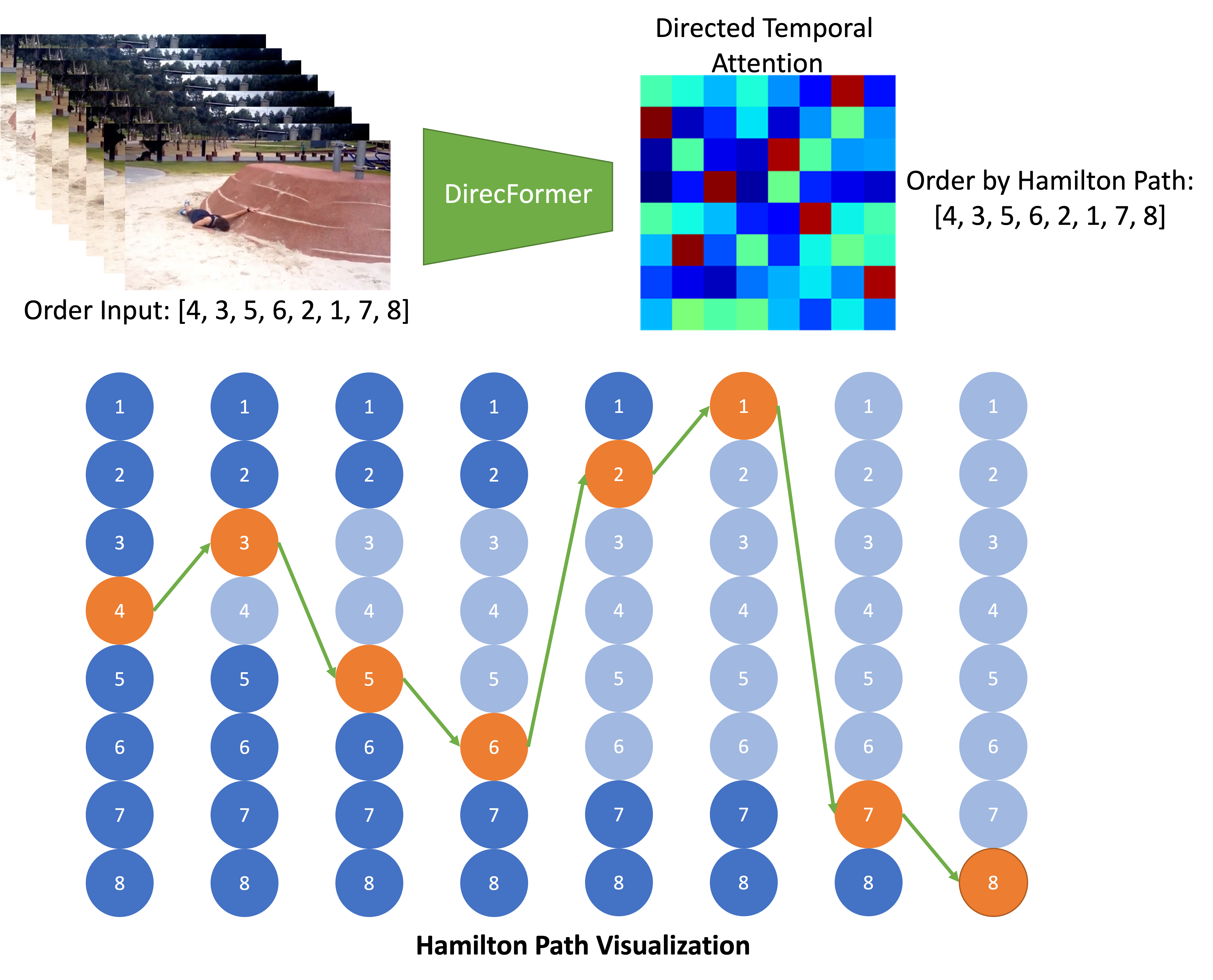}
    \vspace{-4.5mm}
    \caption{Visualization of Order Correction by Finding Hamilton Path On Directed Temporal Attention Map.}
    \vspace{-6.5mm}
    \label{fig:hal_vis}
\end{figure}

\noindent
\textbf{Qualitative Results}
Fig. \ref{fig:hal_vis} illustrates the Directed Attention of our proposed DirecFormer. We use a video on the validation set of Kinetics-400 and extract the attention map. We randomly permute the frames of video along the temporal dimension and correct the order of frames using the Hamilton algorithm. As in Fig. \ref{fig:hal_vis}, we can successfully correct the frame order of the \textit{Parkour} action video.

\section{Conclusions}

In this paper, we have presented a new and simple DirecFormer method with Directed Attention mechanism in Transformer over the temporal and spatial dimensions. The presented Directed Temporal-Spatial Attention not only learns the magnitude of the correlation between frames and tokens, but also exploits the direction of attention. Moreover, the self-supervised guided loss further enhances the directed learning capability of the Directed Temporal Attention.
The intensive ablation study on the Jester dataset has shown the effectiveness of our proposed Directed Attention in both time and space. Furthermore, it has illustrated the impact of the proposed losses used in Directed Temporal Attention learning.
The experiments on two other large-scale datasets, i.e. Something-Something V2 and Kinetics 400, have further confirmed the high accuracy performance of our proposed method.

\noindent
\textbf{Acknowledgement} This work is supported by NSF Data Science, Data Analytics that are Robust and Trusted (DART), Arkansas Biosciences Institute (ABI) Grant Program and NSF WVAR-CRESH Grant.

{\small
\bibliographystyle{ieee_fullname}
\bibliography{arxiv}

\begin{thebibliography}{10}\itemsep=-1pt

\bibitem{vatt}
Hassan Akbari, Liangzhe Yuan, Rui Qian, Wei-Hong Chuang, Shih-Fu Chang, Yin
  Cui, and Boqing Gong.
\newblock Vatt: Transformers for multimodal self-supervised learning from raw
  video, audio and text, 2021.

\bibitem{vivit}
Anurag Arnab, Mostafa Dehghani, Georg Heigold, Chen Sun, Mario Lučić, and
  Cordelia Schmid.
\newblock Vivit: A video vision transformer, 2021.

\bibitem{layer_norm}
Jimmy~Lei Ba, Jamie~Ryan Kiros, and Geoffrey~E. Hinton.
\newblock Layer normalization, 2016.

\bibitem{timesformer}
Gedas Bertasius, Heng Wang, and Lorenzo Torresani.
\newblock Is space-time attention all you need for video understanding?
\newblock In {\em Proceedings of the International Conference on Machine
  Learning (ICML)}, July 2021.

\bibitem{domain_jigsaw_puzzle}
Fabio~Maria Carlucci, Antonio D'Innocente, Silvia Bucci, Barbara Caputo, and
  Tatiana Tommasi.
\newblock Domain generalization by solving jigsaw puzzles, 2019.

\bibitem{kinetic600}
Jo{\~{a}}o Carreira, Eric Noland, Andras Banki{-}Horvath, Chloe Hillier, and
  Andrew Zisserman.
\newblock A short note about kinetics-600.
\newblock {\em CoRR}, abs/1808.01340, 2018.

\bibitem{i3d}
Jo{\~{a}}o Carreira and Andrew Zisserman.
\newblock Quo vadis, action recognition? {A} new model and the kinetics
  dataset.
\newblock In {\em CVPR}, pages 4724--4733. IEEE, 2017.

\bibitem{chen2021deep}
Chun-Fu~Richard Chen, Rameswar Panda, Kandan Ramakrishnan, Rogerio Feris, John
  Cohn, Aude Oliva, and Quanfu Fan.
\newblock Deep analysis of cnn-based spatio-temporal representations for action
  recognition.
\newblock In {\em Proceedings of the IEEE/CVF Conference on Computer Vision and
  Pattern Recognition}, pages 6165--6175, 2021.

\bibitem{imagenet}
Jia Deng, Wei Dong, Richard Socher, Li-Jia Li, Kai Li, and Li Fei-Fei.
\newblock Imagenet: A large-scale hierarchical image database.
\newblock In {\em 2009 IEEE Conference on Computer Vision and Pattern
  Recognition}, pages 248--255, 2009.

\bibitem{bert}
Jacob Devlin, Ming-Wei Chang, Kenton Lee, and Kristina Toutanova.
\newblock Bert: Pre-training of deep bidirectional transformers for language
  understanding, 2019.

\bibitem{poitr_dollar}
P. Dollar, V. Rabaud, Garrison Cottrell, and Serge Belongie.
\newblock Behavior recognition via sparse spatio-temporal features.
\newblock pages 65--72, 2005.

\bibitem{lrcn}
Jeff Donahue, Lisa~Anne Hendricks, Sergio Guadarrama, Marcus Rohrbach,
  Subhashini Venugopalan, Trevor Darrell, and Kate Saenko.
\newblock Long-term recurrent convolutional networks for visual recognition and
  description.
\newblock In {\em CVPR}, pages 2625--2634. IEEE, 2015.

\bibitem{vit}
Alexey Dosovitskiy, Lucas Beyer, Alexander Kolesnikov, Dirk Weissenborn,
  Xiaohua Zhai, Thomas Unterthiner, Mostafa Dehghani, Matthias Minderer, Georg
  Heigold, Sylvain Gelly, Jakob Uszkoreit, and Neil Houlsby.
\newblock An image is worth 16x16 words: Transformers for image recognition at
  scale, 2021.

\bibitem{duong2019dam_ijcv}
Chi~Nhan Duong, Khoa Luu, Kha~Gia Quach, and Tien~D. Bui.
\newblock Deep appearance models: A deep boltzmann machine approach for face
  modeling.
\newblock {\em IJCV}, 2019.

\bibitem{duong2029cvpr_automatic}
Chi~Nhan Duong, Khoa Luu, Kha~Gia Quach, Nghia Nguyen, Eric Patterson, Tien~D.
  Bui, and Ngan Le.
\newblock Automatic face aging in videos via deep reinforcement learning.
\newblock In {\em CVPR}, 2019.

\bibitem{duong2019learning}
Chi~Nhan Duong, Kha~Gia Quach, Khoa Luu, T~Hoang~Ngan Le, Marios Savvides, and
  Tien~D Bui.
\newblock Learning from longitudinal face demonstration—where tractable deep
  modeling meets inverse reinforcement learning.
\newblock {\em IJCV}, 2019.

\bibitem{mvit}
Haoqi Fan, Bo Xiong, Karttikeya Mangalam, Yanghao Li, Zhicheng Yan, Jitendra
  Malik, and Christoph Feichtenhofer.
\newblock Multiscale vision transformers, 2021.

\bibitem{x3d}
Christoph Feichtenhofer.
\newblock X3d: Expanding architectures for efficient video recognition, 2020.

\bibitem{slowfast}
Christoph Feichtenhofer, Haoqi Fan, Jitendra Malik, and Kaiming He.
\newblock Slowfast networks for video recognition.
\newblock In {\em ICCV}, pages 6201--6210. IEEE, 2019.

\bibitem{spatiotemporal_resnet}
Christoph Feichtenhofer, Axel Pinz, and Richard~P. Wildes.
\newblock Spatiotemporal residual networks for video action recognition.
\newblock In Daniel~D. Lee, Masashi Sugiyama, Ulrike von Luxburg, Isabelle
  Guyon, and Roman Garnett, editors, {\em NeurIPS}, pages 3468--3476, 2016.

\bibitem{spatiotemporal_multiplier}
Christoph Feichtenhofer, Axel Pinz, and Richard~P. Wildes.
\newblock Spatiotemporal multiplier networks for video action recognition.
\newblock In {\em CVPR}, pages 7445--7454. IEEE, 2017.

\bibitem{2streams_feichtenhofer}
Christoph Feichtenhofer, Axel Pinz, and Andrew Zisserman.
\newblock Convolutional two-stream network fusion for video action recognition.
\newblock In {\em CVPR}, pages 1933--1941. IEEE, 2016.

\bibitem{ssv2}
Raghav Goyal, Samira~Ebrahimi Kahou, Vincent Michalski, Joanna Materzyńska,
  Susanne Westphal, Heuna Kim, Valentin Haenel, Ingo Fruend, Peter Yianilos,
  Moritz Mueller-Freitag, Florian Hoppe, Christian Thurau, Ingo Bax, and Roland
  Memisevic.
\newblock The "something something" video database for learning and evaluating
  visual common sense, 2017.

\bibitem{ava}
Chunhui Gu, Chen Sun, David~A. Ross, Carl Vondrick, Caroline Pantofaru, Yeqing
  Li, Sudheendra Vijayanarasimhan, George Toderici, Susanna Ricco, Rahul
  Sukthankar, Cordelia Schmid, and Jitendra Malik.
\newblock {AVA:} {A} video dataset of spatio-temporally localized atomic visual
  actions.
\newblock In {\em CVPR}, pages 6047--6056. IEEE, 2018.

\bibitem{resnet}
Kaiming He, Xiangyu Zhang, Shaoqing Ren, and Jian Sun.
\newblock Deep residual learning for image recognition.
\newblock In {\em CVPR}, pages 770--778. IEEE, 2016.

\bibitem{lstm}
Sepp Hochreiter and J{\"{u}}rgen Schmidhuber.
\newblock Long short-term memory.
\newblock {\em Neural Comput.}, 9(8):1735--1780, 1997.

\bibitem{paper_C}
Kai Hu, Jie Shao, Yuan Liu, Bhiksha Raj, Marios Savvides, and Zhiqiang Shen.
\newblock Contrast and order representations for video self-supervised
  learning.
\newblock In {\em 2021 IEEE/CVF International Conference on Computer Vision
  (ICCV)}, pages 7919--7929, 2021.

\bibitem{squeezenet}
Forrest~N. Iandola, Matthew~W. Moskewicz, Khalid Ashraf, Song Han, William~J.
  Dally, and Kurt Keutzer.
\newblock Squeezenet: Alexnet-level accuracy with 50x fewer parameters and
  {\textless}1mb model size.
\newblock {\em CoRR}, abs/1602.07360, 2016.

\bibitem{stm}
Boyuan Jiang, Mengmeng Wang, Weihao Gan, Wei Wu, and Junjie Yan.
\newblock Stm: Spatiotemporal and motion encoding for action recognition, 2019.

\bibitem{sport1m}
Andrej Karpathy, George Toderici, Sanketh Shetty, Thomas Leung, Rahul
  Sukthankar, and Fei{-}Fei Li.
\newblock Large-scale video classification with convolutional neural networks.
\newblock In {\em CVPR}, pages 1725--1732. IEEE, 2014.

\bibitem{kinetic400}
Will Kay, Jo{\~{a}}o Carreira, Karen Simonyan, Brian Zhang, Chloe Hillier,
  Sudheendra Vijayanarasimhan, Fabio Viola, Tim Green, Trevor Back, Paul
  Natsev, Mustafa Suleyman, and Andrew Zisserman.
\newblock The kinetics human action video dataset.
\newblock {\em CoRR}, abs/1705.06950, 2017.

\bibitem{kim2021efficient}
Hanul Kim, Mihir Jain, Jun-Tae Lee, Sungrack Yun, and Fatih Porikli.
\newblock Efficient action recognition via dynamic knowledge propagation.
\newblock In {\em Proceedings of the IEEE/CVF International Conference on
  Computer Vision}, pages 13719--13728, 2021.

\bibitem{cordelia_schimid}
Alexander Kl{\"{a}}ser, Marcin Marszalek, and Cordelia Schmid.
\newblock A spatio-temporal descriptor based on 3d-gradients.
\newblock In Mark Everingham, Chris~J. Needham, and Roberto Fraile, editors,
  {\em BMVC}, pages 1--10. British Machine Vision Association, 2008.

\bibitem{alex_krizhevsky}
Alex Krizhevsky, Ilya Sutskever, and Geoffrey~E. Hinton.
\newblock Imagenet classification with deep convolutional neural networks.
\newblock In Peter~L. Bartlett, Fernando C.~N. Pereira, Christopher J.~C.
  Burges, L{\'{e}}on Bottou, and Kilian~Q. Weinberger, editors, {\em NeurIPS},
  pages 1106--1114, 2012.

\bibitem{kwon2021learning}
Heeseung Kwon, Manjin Kim, Suha Kwak, and Minsu Cho.
\newblock Learning self-similarity in space and time as generalized motion for
  video action recognition.
\newblock In {\em Proceedings of the IEEE/CVF International Conference on
  Computer Vision}, pages 13065--13075, 2021.

\bibitem{ivan_laptev}
Ivan Laptev and Tony Lindeberg.
\newblock Space-time interest points.
\newblock In {\em ICCV}, pages 432--439. IEEE, 2003.

\bibitem{invariant_spatiotemporal_feat}
Quoc~V. Le, Will~Y. Zou, Serena~Y. Yeung, and Andrew~Y. Ng.
\newblock Learning hierarchical invariant spatio-temporal features for action
  recognition with independent subspace analysis.
\newblock In {\em CVPR}, pages 3361--3368. IEEE, 2011.

\bibitem{le2018segmentation}
T.~Hoang~Ngan Le, Kha~Gia Quach, Khoa Luu, Chi~Nhan Duong, and Marios Savvides.
\newblock Reformulating level sets as deep recurrent neural network approach to
  semantic segmentation.
\newblock {\em TIP}, 2018.

\bibitem{li2021memory}
Ce Li, Chunyu Xie, Baochang Zhang, Jungong Han, Xiantong Zhen, and Jie Chen.
\newblock Memory attention networks for skeleton-based action recognition.
\newblock {\em IEEE Transactions on Neural Networks and Learning Systems},
  2021.

\bibitem{li2021else}
Tianjiao Li, Qiuhong Ke, Hossein Rahmani, Rui~En Ho, Henghui Ding, and Jun Liu.
\newblock Else-net: Elastic semantic network for continual action recognition
  from skeleton data.
\newblock In {\em Proceedings of the IEEE/CVF International Conference on
  Computer Vision}, pages 13434--13443, 2021.

\bibitem{li2021trear}
Xiangyu Li, Yonghong Hou, Pichao Wang, Zhimin Gao, Mingliang Xu, and Wanqing
  Li.
\newblock Trear: Transformer-based rgb-d egocentric action recognition.
\newblock {\em IEEE Transactions on Cognitive and Developmental Systems}, 2021.

\bibitem{tsm}
Ji Lin, Chuang Gan, and Song Han.
\newblock {TSM:} temporal shift module for efficient video understanding.
\newblock In {\em ICCV}, pages 7082--7092. IEEE, 2019.

\bibitem{noframe}
Xin Liu, Silvia~L. Pintea, Fatemeh~Karimi Nejadasl, Olaf Booij, and Jan~C. van
  Gemert.
\newblock No frame left behind: Full video action recognition, 2021.

\bibitem{swin}
Ze Liu, Jia Ning, Yue Cao, Yixuan Wei, Zheng Zhang, Stephen Lin, and Han Hu.
\newblock Video swin transformer, 2021.

\bibitem{jester}
Joanna Materzynska, Guillaume Berger, Ingo Bax, and Roland Memisevic.
\newblock The jester dataset: A large-scale video dataset of human gestures.
\newblock {\em 2019 IEEE/CVF International Conference on Computer Vision
  Workshop (ICCVW)}, pages 2874--2882, 2019.

\bibitem{paper_A}
Ishan Misra, C.~Lawrence Zitnick, and Martial Hebert.
\newblock {Shuffle and Learn: Unsupervised Learning using Temporal Order
  Verification}.
\newblock In {\em ECCV}, 2016.

\bibitem{Duong_2017_ICCV}
Chi Nhan~Duong, Kha Gia~Quach, Khoa Luu, Ngan Le, and Marios Savvides.
\newblock Temporal non-volume preserving approach to facial age-progression and
  age-invariant face recognition.
\newblock In {\em ICCV}, Oct 2017.

\bibitem{crosstransformer}
Toby Perrett, Alessandro Masullo, Tilo Burghardt, Majid Mirmehdi, and Dima
  Damen.
\newblock Temporal-relational crosstransformers for few-shot action
  recognition, 2021.

\bibitem{plizzari2021spatial}
Chiara Plizzari, Marco Cannici, and Matteo Matteucci.
\newblock Spatial temporal transformer network for skeleton-based action
  recognition.
\newblock In {\em International Conference on Pattern Recognition}, pages
  694--701. Springer, 2021.

\bibitem{p3d}
Zhaofan Qiu, Ting Yao, and Tao Mei.
\newblock Learning spatio-temporal representation with pseudo-3d residual
  networks.
\newblock In {\em ICCV}, pages 5534--5542. IEEE, 2017.

\bibitem{lgd3d}
Zhaofan Qiu, Ting Yao, Chong-Wah Ngo, Xinmei Tian, and Tao Mei.
\newblock Learning spatio-temporal representation with local and global
  diffusion, 2019.

\bibitem{jason_corso}
Sreemanananth Sadanand and Jason~J. Corso.
\newblock Action bank: {A} high-level representation of activity in video.
\newblock In {\em CVPR}, pages 1234--1241. IEEE, 2012.

\bibitem{2streams_simonyan}
Karen Simonyan and Andrew Zisserman.
\newblock Two-stream convolutional networks for action recognition in videos.
\newblock In Zoubin Ghahramani, Max Welling, Corinna Cortes, Neil~D. Lawrence,
  and Kilian~Q. Weinberger, editors, {\em NeurIPS}, pages 568--576, 2014.

\bibitem{vgg}
Karen Simonyan and Andrew Zisserman.
\newblock Very deep convolutional networks for large-scale image recognition.
\newblock In Yoshua Bengio and Yann LeCun, editors, {\em ICLR}, 2015.

\bibitem{szegedy}
Christian Szegedy, Wei Liu, Yangqing Jia, Pierre Sermanet, Scott~E. Reed,
  Dragomir Anguelov, Dumitru Erhan, Vincent Vanhoucke, and Andrew Rabinovich.
\newblock Going deeper with convolutions.
\newblock In {\em CVPR}, pages 1--9. IEEE, 2015.

\bibitem{inceptionv3}
Christian Szegedy, Vincent Vanhoucke, Sergey Ioffe, Jonathon Shlens, and
  Zbigniew Wojna.
\newblock Rethinking the inception architecture for computer vision.
\newblock In {\em CVPR}, pages 2818--2826. IEEE, 2016.

\bibitem{c3d}
Du Tran, Lubomir~D. Bourdev, Rob Fergus, Lorenzo Torresani, and Manohar Paluri.
\newblock Learning spatiotemporal features with 3d convolutional networks.
\newblock In {\em ICCV}, pages 4489--4497. IEEE, 2015.

\bibitem{ipcsn152}
Du Tran, Heng Wang, Lorenzo Torresani, and Matt Feiszli.
\newblock Video classification with channel-separated convolutional networks,
  2019.

\bibitem{r21d}
Du Tran, Heng Wang, Lorenzo Torresani, Jamie Ray, Yann LeCun, and Manohar
  Paluri.
\newblock A closer look at spatiotemporal convolutions for action recognition.
\newblock In {\em CVPR}, pages 6450--6459. IEEE, 2018.

\bibitem{dat2021bimal_iccv}
Thanh-Dat Truong, Chi~Nhan Duong, Ngan Le, Son~Lam Phung, Chase Rainwater, and
  Khoa Luu.
\newblock Bimal: Bijective maximum likelihood approach to domain adaptation in
  semantic scene segmentation.
\newblock {\em IICV}, 2021.

\bibitem{truong2021fastflow}
Thanh-Dat Truong, Chi~Nhan Duong, Minh-Triet Tran, Ngan Le, and Khoa Luu.
\newblock Fast flow reconstruction via robust invertible n × n convolution.
\newblock {\em Future Internet}, 2021.

\bibitem{iDT}
Heng Wang and Cordelia Schmid.
\newblock Action recognition with improved trajectories.
\newblock In {\em ICCV}, pages 3551--3558. IEEE, 2013.

\bibitem{tdn}
Limin Wang, Zhan Tong, Bin Ji, and Gangshan Wu.
\newblock Tdn: Temporal difference networks for efficient action recognition,
  2021.

\bibitem{tsn}
Limin Wang, Yuanjun Xiong, Zhe Wang, Yu Qiao, Dahua Lin, Xiaoou Tang, and
  Luc~Van Gool.
\newblock Temporal segment networks: Towards good practices for deep action
  recognition.
\newblock In Bastian Leibe, Jiri Matas, Nicu Sebe, and Max Welling, editors,
  {\em ECCV}, volume 9912, pages 20--36. Springer, 2016.

\bibitem{non_local}
Xiaolong Wang, Ross~B. Girshick, Abhinav Gupta, and Kaiming He.
\newblock Non-local neural networks.
\newblock In {\em CVPR}, pages 7794--7803. IEEE, 2018.

\bibitem{actionnet}
Zhengwei Wang, Qi She, and Aljosa Smolic.
\newblock Action-net: Multipath excitation for action recognition, 2021.

\bibitem{geert_willems}
Geert Willems, Tinne Tuytelaars, and Luc~Van Gool.
\newblock An efficient dense and scale-invariant spatio-temporal interest point
  detector.
\newblock In David~A. Forsyth, Philip H.~S. Torr, and Andrew Zisserman,
  editors, {\em ECCV}, volume 5303, pages 650--663. Springer, 2008.

\bibitem{slowfastmg}
Chao-Yuan Wu, Ross Girshick, Kaiming He, Christoph Feichtenhofer, and Philipp
  Kr\"{a}henb\"{u}hl.
\newblock {A Multigrid Method for Efficiently Training Video Models}.
\newblock In {\em {CVPR}}, 2020.

\bibitem{resnext}
Saining Xie, Ross Girshick, Piotr Dollár, Zhuowen Tu, and Kaiming He.
\newblock Aggregated residual transformations for deep neural networks, 2017.

\bibitem{s3d}
Saining Xie, Chen Sun, Jonathan Huang, Zhuowen Tu, and Kevin Murphy.
\newblock Rethinking spatiotemporal feature learning: Speed-accuracy trade-offs
  in video classification.
\newblock In Vittorio Ferrari, Martial Hebert, Cristian Sminchisescu, and Yair
  Weiss, editors, {\em ECCV}, volume 11219, pages 318--335. Springer, 2018.

\bibitem{paper_B}
Dejing Xu, Jun Xiao, Zhou Zhao, Jian Shao, Di Xie, and Yueting Zhuang.
\newblock Self-supervised spatiotemporal learning via video clip order
  prediction.
\newblock In {\em Computer Vision and Pattern Recognition (CVPR)}.

\bibitem{pan}
Can Zhang, Yuexian Zou, Guang Chen, and Lei Gan.
\newblock Pan: Towards fast action recognition via learning persistence of
  appearance, 2020.

\bibitem{trg}
Jingran Zhang, Fumin Shen, Xing Xu, and Heng~Tao Shen.
\newblock Temporal reasoning graph for activity recognition, 2019.

\bibitem{vidtr}
Yanyi Zhang, Xinyu Li, Chunhui Liu, Bing Shuai, Yi Zhu, Biagio Brattoli, Hao
  Chen, Ivan Marsic, and Joseph Tighe.
\newblock Vidtr: Video transformer without convolutions, 2021.

\bibitem{zhang2021stst}
Yuhan Zhang, Bo Wu, Wen Li, Lixin Duan, and Chuang Gan.
\newblock Stst: Spatial-temporal specialized transformer for skeleton-based
  action recognition.
\newblock In {\em Proceedings of the 29th ACM International Conference on
  Multimedia}, pages 3229--3237, 2021.

\bibitem{trn}
Bolei Zhou, Alex Andonian, Aude Oliva, and Antonio Torralba.
\newblock Temporal relational reasoning in videos.
\newblock In Vittorio Ferrari, Martial Hebert, Cristian Sminchisescu, and Yair
  Weiss, editors, {\em ECCV}, volume 11205 of {\em Lecture Notes in Computer
  Science}, pages 831--846. Springer, 2018.

\bibitem{msnet}
Xiaoyu Zhu, Junwei Liang, and Alexander Hauptmann.
\newblock Msnet: A multilevel instance segmentation network for natural
  disaster damage assessment in aerial videos, 2020.

\end{thebibliography}
}

\end{document}